\documentclass[]{fairmeta}

\usepackage{amsmath,amsfonts,bm}

\def\eqref#1{equation~\ref{#1}}

\def\1{\bm{1}}

\DeclareMathAlphabet{\mathsfit}{\encodingdefault}{\sfdefault}{m}{sl}
\SetMathAlphabet{\mathsfit}{bold}{\encodingdefault}{\sfdefault}{bx}{n}

\usepackage{url}
\usepackage{amsmath,amsthm}
\usepackage{amssymb}
\usepackage{mathtools}
\usepackage{array}
\usepackage{flafter,adjustbox,float}

\hypersetup{
 pdftitle={Delayed Effects of Temporary Weight Decay in Muon Pretraining},
 pdfauthor={Md Rifat Arefin, Ravid Shwartz-Ziv, Ernie Chang, Chinnadhurai Sankar, Rylan Conway, Aristide Baratin, Irina Rish, Adithya Sagar, Patrick Huber},
 pdfkeywords={Muon, weight decay, language model pretraining, optimization dynamics}
}

\DeclareMathOperator{\tr}{tr}
\DeclareMathOperator{\OffDiag}{OffDiag}

\newcommand{\WV}{W_V}
\newcommand{\WO}{W_O}
\newcommand{\Wdown}{W_{\mathrm{down}}}
\newcommand{\Wup}{W_{\mathrm{up}}}
\newcommand{\WQ}{W_Q}
\newcommand{\WK}{W_K}

\newcommand{\reluq}{\mathrm{ReLU}^2}

\newcommand{\researchquestion}[1]{%
 \begin{center}
  \setlength{\fboxsep}{7pt}%
  \fcolorbox{black!35}{metabg}{%
   \parbox{0.91\linewidth}{\centering\textbf{Central question.} #1}%
  }
 \end{center}
}

\title{Delayed Effects of Temporary Weight Decay in Muon Pretraining}

\author[1,2,3,*]{Md~Rifat~Arefin}
\author[1]{Ravid~Shwartz-Ziv}
\author[1]{Ernie~Chang}
\author[1]{Chinnadhurai~Sankar}
\author[1]{Rylan~Conway}
\author[2,3,4]{Aristide~Baratin}
\author[2,3]{Irina~Rish}
\author[1]{Adithya~Sagar}
\author[1]{Patrick~Huber}

\affiliation[1]{Meta}
\affiliation[2]{Mila-Quebec AI Institute}
\affiliation[3]{Université de Montréal}
\affiliation[4]{Samsung SAIL Montreal}

\contribution[*]{Work done during an internship at Meta}

\abstract{A temporary optimization intervention can increase validation loss when it ends yet reduce it at the final training budget. We demonstrate this with temporary weight decay in Muon-trained decoder-only Transformers with 162M, 405M, 1.4B, and 3.4B parameters. For models up to 1.4B parameters, moving a fixed-duration decay window from early to late training reverses its final effect. At 162M, changing window duration alters final loss even at approximately matched total shrinkage. Most strikingly, a schedule selected at 162M and transferred without retuning to 1.4B is 0.26 nats worse than no decay when removed, overtakes it only after 73\% of training, and ultimately finishes 0.028 nats better while also outperforming the best sampled constant-decay baseline. Early decay leaves a persistent weight-norm deficit, which keeps relative steps larger after the intervention ends. An approximate norm-dynamics recurrence captures the slow convergence of these relative steps, and fixed-norm replays of the measured trajectories reproduce the performance ordering and a similar loss difference. At 162M and 405M, norm-reset experiments show that smaller weight norms at the end of the decay window explain most of the final loss improvement. A separate Gram intervention exhibits related post-removal norm and relative-step dynamics. These results show that temporary optimization schedules can have delayed effects long after removal and should therefore be evaluated at the intended final training budget, not only at removal.
}

\date{November 12, 2026}
\correspondence{Md Rifat Arefin at \email{rifat.arefin@mila.quebec}}

\begin{document}
\raggedbottom
\renewcommand{\topfraction}{0.9}\renewcommand{\bottomfraction}{0.5}
\renewcommand{\textfraction}{0.1}\renewcommand{\floatpagefraction}{0.85}
\setcounter{topnumber}{1}\setcounter{bottomnumber}{1}\setcounter{totalnumber}{3}

\maketitle

\section{Introduction}
\label{sec:introduction}

\begin{figure}[!t]
 \centering
 \includegraphics[width=0.84\linewidth]{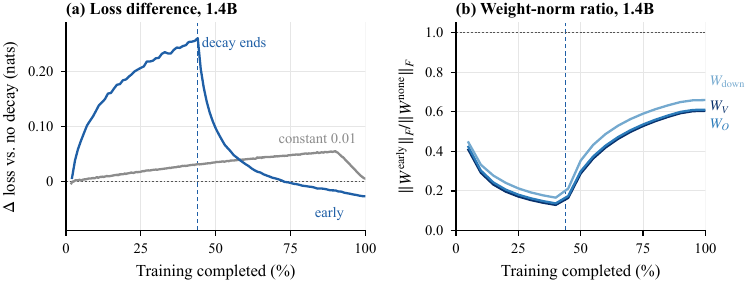}
 \caption{\textbf{A temporary intervention can be harmful at removal and beneficial much later.} 1.4B, FineWeb100B, 28,000
 updates. Decay ($\lambda=0.15$) is applied for the first 44\% of training;  the schedule was selected
 at 162M and transferred without retuning.  (a) Validation-loss difference from no decay; constant decay $0.01$ is shown for comparison. Temporary decay is 0.26 nats worse at removal, crosses no decay at $73\%$ of training, and finishes 0.0275 nats better. (b) Layer-averaged weight norms relative to no decay. The dashed line marks removal; the norm deficit persists afterward.
 }
 \label{fig:first}
\end{figure}

A temporary optimization intervention can increase validation loss when it ends and reduce it much later, even after the intervention has been removed. Figure~\ref{fig:first} shows an extreme example. In a 1.4B-parameter Transformer~\citep{vaswani2017attention} trained with Muon~\citep{jordan2024muon}, a temporary weight-decay schedule leaves validation loss $0.26$ nats worse than no decay when the schedule ends. Both runs then continue with the same explicit update rule. Nevertheless, the treated run gradually recovers, overtakes the no-decay baseline only after 73$\%$ of training, and finishes $0.0275$ nats better. Performance at the endpoint of an intervention can therefore be a poor predictor of its eventual effect.

Temporary changes during early training are known to have lasting effects on the learned solution \citep{achille2019critical,frankle2020early}, and prior work has shown that regularization can matter disproportionately during early learning \citep{golatkar2019time}. We study a different aspect of this phenomenon: the effect of a temporary intervention can continue to evolve after the intervention has ended, to the point that its sign or ranking changes at later checkpoints.

This creates a basic problem for evaluating temporary optimization schedules. Such schedules are usually specified by an update rule and the interval over which it is applied. But when the intervention ends, the treated and reference runs generally enter the subsequent phase with different parameters and optimizer states. Their loss difference can therefore continue to evolve even though their explicit update rules are again identical. Evaluating an intervention when it is removed can consequently select a schedule that is inferior at the intended final training budget---or reject one that will eventually be beneficial.

We study this phenomenon using temporary weight decay in Muon-trained decoder-only Transformers. Weight decay provides a particularly controlled setting: turning it off cleanly separates its direct shrinkage from the training state it leaves behind. Muon also provides a useful connection between weight norms and subsequent optimization. Its matrix updates have approximately controlled magnitude, so for a given projected update norm, reducing the weight norm increases the update size relative to the current weights \citep{jordan2024muon,kosson2024rotational,wen2026hyperball,liu2026effective}. Temporary decay can therefore alter subsequent optimization even after its coefficient has returned to zero.

We ask two related questions:

\researchquestion{How should a temporary optimization intervention be evaluated, 
and what aspects of the training state carry its effect after the intervention is removed?}

We address the first question by varying when decay starts, how long it lasts, and when its effect is evaluated.
Across models from 162M to 1.4B parameters, moving a fixed-duration decay window from early to late training reverses its final effect: early windows improve final loss, whereas a window beginning at $60\%$ of training hurts. At 405M, even among beneficial early schedules, the ranking changes depending on whether they are compared at a common final training budget or after the same amount of post-removal training. At 162M, changing window duration also changes final loss substantially even when total multiplicative shrinkage is approximately matched. These results separate three notions that are often conflated: when an intervention acts, how long the model trains after it is removed, and the final budget at which it is evaluated.

We address the second question by examining what remains different after decay is removed. Early decay leaves a persistent weight-norm deficit, which keeps relative steps larger than in the no-decay run long after both models resume the same update rule. An approximate recurrence accounts for the slow convergence of these relative-step trajectories. An opposite perturbation---boosting the learning rate early---raises weight norms and leaves relative steps smaller after the boost ends, as predicted by the same account. More directly, fixed-norm replays using the measured relative-step trajectories reproduce the performance ordering and a similar loss difference; swapping trajectories at the end of the decay window shows that most of the replay advantage arises from the altered steps after decay has ended. These experiments explain how an early intervention can continue to modify optimization after removal, although they do not establish that norms are the only persistent state or explain why a particular relative-step trajectory is beneficial.

Together, these experiments characterize both the evaluation consequences of delayed optimization effects and one mechanism by which an early intervention can remain active through the state it leaves behind.

\section{Related Work}
\label{sec:related-work}

\paragraph{Early-training interventions.} Temporary changes during early optimization can have lasting effects on the learned solution.
\citet{achille2019critical} and \citet{frankle2020early} study critical early phases in neural-network training, showing that perturbations applied near initialization can influence later optimization and the final solution. Most directly related, \citet{golatkar2019time} show in image
classification that weight decay and data augmentation matter disproportionately during early learning and comparatively little near convergence.
Our focus is different: we study what happens after a temporary intervention is removed. In our setting, the loss difference can continue to evolve for a large fraction of training, reverse sign, and change the ranking of schedules even though the treated and reference runs again use the same explicit update rule.

\paragraph{Weight norms and effective step size.}
Several recent works characterize the relation between weight decay, parameter norms, and effective learning rate.
\citet{kosson2024rotational} analyze rotational equilibrium under weight decay; \citet{wen2026hyperball} derive related equilibrium behavior for Muon and introduce Hyperball, which controls weight and update norms; 
\citet{liu2026effective} connect effective learning rate to language-model loss dynamics;
\citet{kosson2024warmup} analyze the large relative steps produced by weights below equilibrium.
These works characterize persistent coefficients, equilibrium behavior, or explicit norm control. We instead study the transient after temporary decay is removed: the weights can remain far below the no-decay norm trajectory, causing relative step sizes to remain different long after both runs resume the same explicit update rule.

\paragraph{Muon-specific analyses.} Recent work has characterized Muon through spectral-norm constraints and row-norm control \citep{chen2025muon, lion2026muown}. Our
experiments are complementary: rather than modifying Muon's norm control directly, we ask how an earlier temporary intervention changes the optimization trajectory that Muon follows afterward. The resulting norm deficit provides a concrete state variable through which the effect of a removed intervention can persist.

\section{Experimental Setting and Evaluation}
\label{sec:design}

We train decoder-only Transformers with 162M, 405M, 1.4B, and 3.4B parameters. The main
configuration uses Muon, a matrix optimizer that orthogonalizes momentum updates, with
squared-ReLU ($\reluq$) MLPs, a context length of 1,024, and 1,024
sequences per update. The learning rate is 0.01 for the first 90\% of training and decreases
linearly to zero over the final 10\%. The default configuration has no initial warmup; 
Section~\ref{sec:warmup} tests 10\% linear warmup and shows that the delayed crossover persists. The compute-optimal
budget is about 20 tokens per parameter \citep{hoffmann2022training}. Matched intervention
and reference runs share initialization, data order, and learning-rate schedule.

Main timing and scale experiments use FineWeb100B,  a 100B-token sample of FineWeb, while the replicated mechanism experiments use FineWeb-Edu \citep{penedo2024the}. 
Table~\ref{tab:map} in Appendix~\ref{app:setup} gives the dataset and reference for each experiment group. Losses are compared only within matched groups. 

\paragraph{Temporary decay.}
For each  hidden-layer weight matrix, decoupled weight decay \citep{loshchilov2019decoupled} gives
\begin{equation}
 W_{t+1}=(1-\eta_t\lambda_t)W_t-\eta_t U_t,
 \qquad \lambda_t=\lambda\mathbf{1}_{\{s\leq t<e\}},
 \label{eq:update}
\end{equation}
where $U_t$ is the Muon update before multiplication by the learning rate $\eta_t$, and
$[s,e)$ is the intervention window. Decay is zero outside this window unless a constant
background coefficient is stated. Embeddings and the output head receive no decay.

\paragraph{Evaluation and references.}
All losses are next-token cross-entropy on held-out validation tokens, in nats per token
(Appendix~\ref{app:setup}). We report
$\Delta_t=\mathcal L_{\rm intervention}(t)-\mathcal L_{\rm reference}(t)$ at update $t$;
the final difference is $\Delta_T$ at the last update $T$, and negative values indicate an improvement. We call the end of
the window \emph{removal}. In the timing experiments, \emph{early} windows start within the first
20\% of training and the \emph{late} window starts at 60\%. The \emph{best sampled constant} is the best of the constant
coefficients we tried for that scale and training setting, and the \emph{best reference} is the better of this baseline and
no decay. In the main timing and scale comparisons, claims that a schedule helps are made against both. Other experiment groups use the references specified in Appendix~\ref{app:setup}. We use no decay
alone when asking what the window itself changes (Section~\ref{sec:explanation}), because
constant decay adds norm dynamics of its own.

\section{Temporary Decay Has Delayed Effects}
\label{sec:early-benefits}
\label{sec:when}

Having established the delayed crossover in Figure~\ref{fig:first}, we now separate three properties of a temporary schedule: when the window starts, how long it lasts, and when its effect is evaluated. These factors need not agree in their ranking of schedules because they determine both the state at removal and the subsequent optimization trajectory.

\subsection{A Short Early Window Improves Final Loss}
\label{sec:both}

We first verify the effect in a replicated comparison. Across five paired 162M runs, a 320-update early-decay window improves final loss by 0.0172 nats over no decay and by 0.0074 nats over the best sampled constant (Appendix~\ref{app:schedules},  Table~\ref{tab:seeds}).

\subsection{Start Time Changes the Sign; Evaluation Time Changes the Ranking}
\label{sec:start-time}

\begin{figure}[tb]
 \centering
 \includegraphics[width=0.84\linewidth]{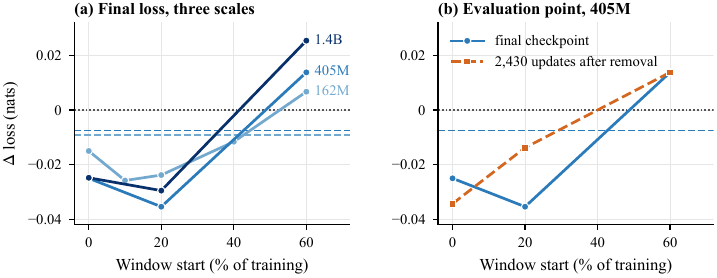}
 \caption{\textbf{Start time changes the sign; evaluation horizon changes the ranking.}
 (a) Final loss difference for 10\% windows by start time; dashed lines show the best sampled constant. (b) At 405M, the ranking of early starts reverses when evaluated after equal post-removal training.}
 \label{fig:timing}
\end{figure}

At all three scales, a 10\% window starting at initialization beats both references, whereas the same-duration, same-coefficient window starting at 60\% is worse than both
(Figure~\ref{fig:timing}(a); Appendix~\ref{app:schedules}).
Within each scale these windows lie in the constant-learning-rate phase and apply the same multiplicative shrinkage, so the sign reversal is a timing effect rather than a difference in total shrinkage.

Comparing at the final checkpoint fixes total training, whereas comparing a fixed number of updates after removal fixes post-removal training. At 405M these two evaluations reverse the ranking of the early schedules (Figure~\ref{fig:timing}(b)): the $20\%$-start window is better at the common final checkpoint, while the initialization-start window is better after equal post-removal training. For the $0\%$ and $20\%$ starts, the equal-continuation checkpoints remain in the constant-learning-rate phase, so the reversal is not caused by the final learning-rate decay; the $60\%$ start is less directly comparable because its continuation enters that phase.  Thus, the preferred early start depends on the evaluation horizon, although early starts help and the late start hurts under both comparisons.

\subsection{Duration Changes Outcomes beyond Total Shrinkage}
\label{sec:duration-grid}

\begin{figure}[tb]
 \centering
 \includegraphics[width=0.8\linewidth]{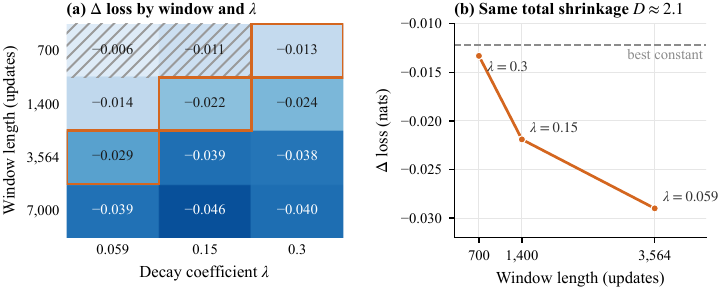}
 \caption{\textbf{Duration affects loss beyond total shrinkage.}
 162M, FineWeb100B, 16,000 updates. (a) Final loss difference across duration and decay coefficient; hatching marks schedules that do not beat the best sampled constant.
 (b) At matched $D\approx 2.1$, longer, weaker
 decay performs better.}
 \label{fig:duration}
\end{figure}

Across our grid, final loss is more sensitive to window duration than to the decay coefficient. At fixed coefficient, varying duration changes the final improvement by 0.026--0.035 nats; at fixed duration, varying the coefficient changes it by only 0.007--0.010 nats.  We train 162M models for 16,000
updates, five times the compute-optimal budget, in a $4\times3$ grid of durations and
coefficients, with every window starting at initialization. At each coefficient, longer windows
give lower loss (Figure~\ref{fig:duration}(a)).  The best sampled constant, $0.05$, improves on no decay by 0.0122. Ten of the
twelve cells beat it; the two that do not are the shortest windows with the two smallest
coefficients.

Longer windows also apply more decay in total. To separate the two effects, we compare
schedules with approximately equal total log-shrinkage,
\begin{equation}
 D=-\sum_t\log(1-\eta_t\lambda_t).
 \label{eq:shrink}
\end{equation}
For three schedules with $D\approx2.1$, the magnitude of the final improvement more than doubles as the window lengthens from
700 to 3,564 updates (Figure~\ref{fig:duration}(b)). Matching $D$ fixes
how much the initial weights are scaled down, but not how later updates are scaled
(Appendix~\ref{app:derivations}).

The best schedule in this grid,  $\lambda=0.15$ over the first 44\% of training, improves on the best sampled
constant by 0.0335. We select this schedule here, at 162M, and transfer it without retuning to the 1.4B experiment in Figure~\ref{fig:first} and Section~\ref{sec:transfer_schedule}.  It is also the longest window in this grid, so these runs do not locate the
optimal duration. Separate fixed-coefficient sweeps at the compute-optimal budget find an
interior optimum at both 162M and 405M, beyond which longer windows hurt
(Appendix~\ref{app:schedules}).

Together, these experiments show that a temporary schedule cannot be ranked from its removal-time loss or total shrinkage alone. Start time, duration, and evaluation horizon can all change the comparison, because they alter either the state at removal or the amount of subsequent optimization. We next ask which component of that state can preserve the effect of an earlier decay window.

\section{A Persistent Norm Deficit Changes Later Relative Steps}
\label{sec:explanation}
\label{sec:relative-step}

Once decay is removed, the explicit update rule is again the same as in the no-decay run. Any continuing effect must therefore be carried by the parameters or optimizer state created during the intervention. Temporary decay can change weight directions, weight norms, and momentum. We focus on norms because they directly affect the size of a Muon update relative to the current weights. We first show that early decay leaves a long-lived norm deficit and correspondingly larger relative steps; we then test the relative-step trajectory directly with fixed-norm replays, before analyzing why the difference converges only slowly. This analysis explains how an early intervention can continue to modify optimization after removal, but does not establish that norms are the only persistent state or explain why a particular relative-step trajectory lowers loss.

\subsection{Early decay creates a persistent norm deficit}

\begin{figure}[!t]
 \centering
 \includegraphics[width=0.8\linewidth]{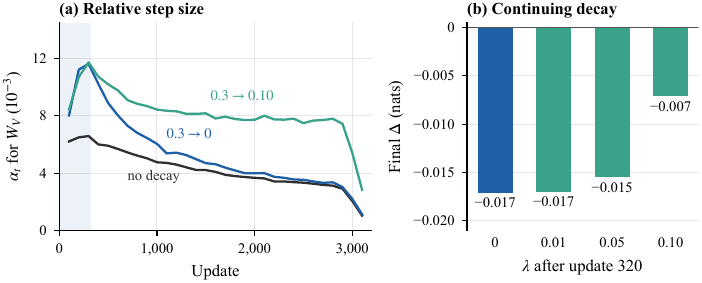}
 \caption{\textbf{Early decay increases relative step size, but larger relative steps do not
 always improve loss.} (a) Layer-averaged relative step size $\alpha_t$ for $\WV$; the shaded
 band is the decay window. (b) Final loss differences from no decay after using $\lambda=0.3$
 for the first 320 updates and then the indicated coefficient. 162M, FineWeb-Edu, 3,200 updates.}
 \label{fig:norm-controls}
\end{figure}

In the transferred 1.4B run of Figure~\ref{fig:first}(b), norms fall to 12--16\% of their no-decay values at removal and remain only 60--66\% as large at the final checkpoint. The same qualitative deficit persists in the 162M mechanism runs.

\subsection{The norm deficit produces persistently larger relative steps}
\label{sec:normdeficit}

For the weight change $\Delta W_t=W_{t+1}-W_t$, we define the relative step size as the part of
the change perpendicular to $W_t$, divided by the norm of $W_t$:
\begin{equation}
 \alpha_t=\frac{\|P_{W_t}^{\perp}\Delta W_t\|_F}{\|W_t\|_F}
 =\eta_t\frac{\|P_{W_t}^{\perp}U_t\|_F}{\|W_t\|_F},
 \label{eq:angular}
\end{equation}
where $P_W^{\perp}X=X-\langle W,X\rangle_F W/\|W\|_F^2$. This projection removes the decay
term, which is parallel to $W$. For a fixed projected update norm, $\alpha_t$ is inversely
proportional to $\|W_t\|_F$, as in norm-based analyses of the effective learning rate
\citep{kosson2024rotational,wen2026hyperball,liu2026effective}.

At 162M, early decay lowers norms and increases $\alpha_t$. After decay is removed, the relative steps approach the no-decay trajectory but remain larger until the end of training
(Figure~\ref{fig:norm-controls}(a)). The norm deficit is not the whole story: at update 2,800, the inverse norm ratio is 1.17 whereas the relative-step ratio is 1.07, showing that changes in the projected update norm partly offset the effect of the smaller weights.

An early learning-rate boost provides the opposite perturbation. It raises weight norms and, as predicted by the same inverse relationship, leaves subsequent relative steps smaller than in the no-decay run. All three tested boosts finish with higher loss (Appendix~\ref{app:extra}). 

\subsection{Replaying the relative-step trajectory reproduces the gain}

\begin{figure}[!t]
 \centering
 \includegraphics[width=\linewidth]{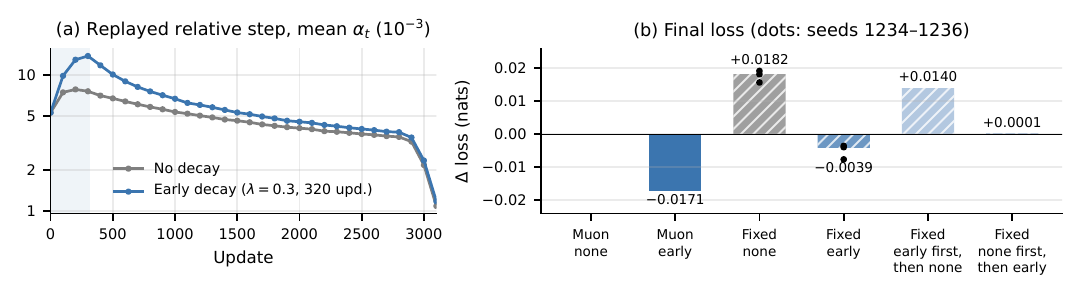}
 \caption{\textbf{Fixed-norm replay reproduces the post-removal gain.} 162M, FineWeb-Edu, 3,200 updates. (a) Relative-step trajectories measured from the source Muon runs and replayed at fixed weight norms. (b) Final loss relative to Muon without decay; dots show three replay seeds and the last two bars switch trajectory at update 320.}
 \label{fig:replay}
\end{figure}

The correlation between smaller norms and larger relative steps does not establish that the altered relative-step trajectory matters for loss. To test this directly, we remove norm variation altogether.
Using a variant of Hyperball \citep{wen2026hyperball}, we hold every weight norm at its initial value and prescribe the relative-step trajectory $\alpha_t$ measured from either the no-decay or early-decay Muon run, averaged over the six matrix families (Figure~\ref{fig:replay}(a)).

Across three replay seeds, the trajectory measured from the early-decay run improves final loss by 0.022 to 0.023
nats relative to the no-decay trajectory, compared with a 0.0171-nat difference between the Muon runs from which the trajectories were taken (Figure~\ref{fig:replay}(b)). Thus, with norms fixed, the measured relative-step trajectories reproduce both the ordering and a similar magnitude of the original loss difference. 

A trajectory-swap experiment localizes where this difference arises. Using no-decay steps during the first 320 updates and early-decay steps afterward accounts for 0.0181 of the 0.0221-nat replay difference. Conversely, replaying only the early-decay trajectory during the intervention and then returning to ordinary Muon improves loss by only 0.0003 nats. Most of the replay advantage therefore comes from the altered relative steps after weight decay has already been removed (Figure~\ref{fig:replay}(b)).

The replay uses a modified optimizer and averages $\alpha_t$ over matrix families, so it does not show that relative steps are the only relevant state. It does show that the observed post-removal relative-step trajectory is sufficient, in this controlled setting, to reproduce the performance ordering and approximate loss difference. Resetting block norms at removal removes or recovers most of the early-decay advantage at 162M and 405M (Appendix~\ref{app:norm-reset}).

\subsection{Why do the trajectories converge so slowly?}

We next ask why the relative-step difference created by a short intervention decays so slowly.

Let $a_t=\|\Delta W_t\|_F/\|W_t\|_F$ and let $c_t$ be the cosine between $W_t$ and $\Delta W_t$,
so $\alpha_t=a_t\sqrt{1-c_t^2}$. From update 400 on, $|c_t|<0.20$, so 
$\alpha_t/a_t>\sqrt{0.96}\approx0.9798$. Expanding $\|W_t+\Delta W_t\|_F^2$ and approximating consecutive Muon
updates as equal in norm gives
\begin{equation}
 \frac{1}{a_{t+1}^2}-\frac{1}{a_t^2}\approx1+2\kappa_t,\qquad
 \kappa_t=\frac{\langle W_t,\Delta W_t\rangle_F}{\|\Delta W_t\|_F^2}.
 \label{eq:inverse-step-main}
\end{equation}
If the recurrence increments are similar in the early-decay and no-decay runs, their offset in $1/a_t^2$ remains approximately constant. Meanwhile $1/a_t^2$ itself grows. The fixed offset therefore becomes a progressively smaller fraction of the total, causing the ratio of relative steps to converge slowly rather than disappear immediately. Appendix~\ref{app:extra}  derives the recurrence from the exact norm identity and reports the corresponding checks across matrix families.

At 162M, $\kappa_t$ is similar between the two runs, and the measured growth of $1/a_t^2$ agrees closely with the recurrence for four
matrix families. From 400 to 2,800, $1/\alpha_t^2$ grows roughly fourfold whereas the gap between the runs changes much less, and the relative-step ratio falls only from about 1.6--1.8 to about 1.1. Appendix~\ref{app:extra} quantifies these deviations and shows the per-family diagnostics. 

The early learning-rate boosts introduced in Section~\ref{sec:normdeficit} provide an opposite-sign check of this account: they raise the norms and produce persistent reductions in later relative steps, with the measured offsets behaving as predicted by the same recurrence
(Appendix~\ref{app:extra}). 

At 1.4B, we do not have the corresponding relative-step logs for the no-decay run, so we cannot test the recurrence as directly. We can nevertheless track the norm deficit itself. From update 14,000 to 25,200,  the layer-averaged squared-norm gap between the no-decay and early-decay runs grows by $39\%$ for $\WV$ and $\WO$ and $17\%$ for $\Wdown$ while the squared norms in the no-decay run roughly double. Thus, the absolute deficit does not remain fixed, but it becomes progressively smaller relative to the growing weight norms, consistent with the slow recovery seen at 162M.

\paragraph{What norms do not explain.}
The norm account explains why an early intervention can continue to alter optimization after removal, but not which relative-step trajectory is beneficial. Larger relative steps are not universally better. Continuing with decay coefficient $0.10$ after the window
keeps relative steps larger than stopping decay, yet yields a smaller final improvement (0.0071  
versus 0.0171 nats; Figure~\ref{fig:norm-controls}(b)). Likewise, a window starting at 60\% of training shrinks
the weights by the same factor as one starting at initialization but hurts final loss
(Figure~\ref{fig:timing}(a)). Thus, persistent norm and relative-step differences explain the delayed influence of the intervention, not why early and late trajectories have opposite effects.

\section{Scale Transfer and Robustness}
\label{sec:transfer}

We next test whether the delayed effect persists across model scale and training conditions. We first transfer a schedule selected at 162M directly to 1.4B without retuning, then compare shorter windows across scale and test robustness to longer training, warmup, architecture, and optimizer choice.

\subsection{A schedule selected at 162M transfers to 1.4B}
\label{sec:transfer_schedule}

We transfer the best schedule from the 162M duration sweep---$\lambda=0.15$ for the first $44\%$ of training---to a 1.4B model without retuning. It finishes 0.0275 nats
ahead of no decay and 0.0264 ahead of the best sampled constant-decay baseline; as Figure~\ref{fig:first} shows, this advantage emerges entirely after removal. The best schedule sampled directly at 1.4B improves on the transferred schedule by only \(0.0020\) nats.

\subsection{The effect persists across scale}

A 10\% decay window starting at initialization beats both no decay and the best sampled constant  at every scale
(Table~\ref{tab:scale-practical}). At 405M, the first-10\% window improves over no decay by \(0.0247\)--\(0.0262\) nats across three matched seeds, and over constant \(0.02\) by \(0.0172\)--\(0.0194\) nats. More broadly, every tested constant coefficient improves on no decay at 162M and 405M, whereas at 1.4B only the two smallest tested coefficients do so, and only slightly. These comparisons show that the benefit of an early window persists across the tested scales, but not that its magnitude follows a scaling law: the fixed \(10\%\) schedule is also farther from the best sampled duration at 162M than at the larger scales. Appendix~\ref{app:schedules} reports the complete constant and schedule sweep. At 3.4B, every tested early schedule improves loss and mean accuracy (Appendix~\ref{app:b34}). 

\begin{table}[tb]
 \centering
 \small
 \caption{\textbf{Early decay across scale.} FineWeb100B,
 compute-optimal budgets, no warmup. ``Best const.'' is the best sampled constant coefficient; full sweeps are in
 Appendix~\ref{app:schedules} and the 3.4B configurations are in Appendix~\ref{app:b34}.}
 \label{tab:scale-practical}
 \vspace{0.55em}
 \begin{adjustbox}{max width=\linewidth}
 \begin{tabular}{lrlrrrr}
  \toprule
  Scale & Updates & Early schedule ($\lambda$, window) & Early loss & No decay & Best const. & vs.\ best ref. \\
  \midrule
  162M & 3,200  & 0.3, 10\% & 3.3710 & 3.3860 & 3.3768 (0.05) & $-0.0058$ \\
  405M & 8,100  & 0.3, 10\% & 3.0287 & 3.0537 & 3.0462 (0.02) & $-0.0175$ \\
  1.4B & 28,000 & 0.2, 10\% & 2.6506 & 2.6754 & 2.6743 (0.003) & $-0.0237$ \\
  1.4B & 28,000 & 0.15, 44\% (transferred) & 2.6479 & 2.6754 & 2.6743 (0.003) & $-0.0264$ \\
  3.4B & 66,000 & 0.2, 10\% & 2.4720 & 2.4832 & 2.4843 (0.001) & $-0.0112$ \\
  3.4B & 66,000 & 0.15, 44\% (transferred) & 2.4782 & 2.4832 & 2.4843 (0.001) & $-0.0050$ \\
  3.4B & 66,000 & 0.15, 12,320 updates (transferred) & 2.4638 & 2.4832 & 2.4843 (0.001) & $-0.0194$ \\
  \bottomrule
 \end{tabular}
 \end{adjustbox}
\end{table}

\subsection{Robustness to training conditions}

\label{sec:warmup}

The effect is not limited to the compute-optimal, no-warmup configuration. At 162M trained for 64,000 updates---$20\times$ the compute-optimal budget---a first-10\% window still beats the best sampled constant by 0.0180 nats (Appendix~\ref{app:horizon}). With 10\% linear warmup, a decay window confined to warmup ties the best sampled constant at 162M and beats it at 405M and 1.4B; extending decay beyond warmup restores a clear advantage at 162M and 405M. Importantly, warmup does not eliminate the delayed crossover: in a 16,000-update 162M run, temporary decay is 0.0237 nats worse than no decay at update 8,000 but 0.0373 nats better by update 14,400
(Appendix~\ref{app:scope}). Early decay also improves loss with SwiGLU MLPs and with AdamW at 162M, but not with AdamW at 405M (Appendix~\ref{app:regime}). Thus, the phenomenon is robust to several training variations but is not optimizer-independent. 

Across four data domains, temporary decay improves on constant \(0.01\) both at removal and at the final checkpoint, but the domain ranking changes between the two evaluations; Appendix~\ref{app:domains} reports these comparisons and exploratory corpus-level analyses.

\section{A Second Norm-Changing Intervention Shows Related Dynamics}
\label{sec:gram}

The preceding mechanism could be specific to the multiplicative form of weight decay. We therefore test a qualitatively different intervention that also changes weight norms. During an early window, we add the off-diagonal Gram penalty
\begin{equation}
 \mathcal R_t=\mu\mathbf{1}_{\{0\leq t<e\}}
 \sum_{\ell=1}^{L}\sum_{W\in\{\WV,\WO,\Wdown\}}E(W_\ell),
 \qquad E(W)=\|\OffDiag(W^\top W)\|_F^2.
 \label{eq:gram}
\end{equation}
Muon orthogonalizes the combined task and penalty gradient, so the applied update is not the raw penalty gradient (Appendix~\ref{app:setup}).

The Gram penalty exhibits the same qualitative timing dependence as temporary decay. At 162M, 10\% windows starting at 0\%, 20\%, and 60\% of training change final loss by -0.0147, -0.0080, and +0.0209 nats, respectively. Across five paired 162M runs, an early Gram penalty improves final loss by 0.0142  nats relative to no regularization.

The raw penalty gradient contains a radial component that changes the weight norm and a tangential component that changes only direction to first order. Radial--tangential and state-copy controls indicate that norm changes account for much of the improvement: when applied after the Muon update, the radial component nearly matches the full penalty while the tangential component has little effect, and copying column norms recovers more of the advantage than copying normalized column correlations (Appendix~\ref{app:derivations}). The penalty also leaves larger post-removal relative steps primarily in the matrices it targets (Appendix~\ref{app:extra}).

Early Gram regularization also improves loss at 162M and 405M across both tested datasets, while 1.4B requires retuning of its strength (Appendix~\ref{app:gram}).

Together, these results provide evidence  that the post-removal norm and relative-step dynamics of Section~\ref{sec:explanation} are not unique to multiplicative weight decay, without implying that norm reduction is generally beneficial.

\section{Discussion and Conclusion}
\label{sec:limitations}

The endpoint of a temporary optimization intervention need not be the endpoint of its effect. In Muon pretraining, temporary decay can be worse at removal yet better at the final budget; both the sign and ranking of schedules can change after removal. This behavior persists across the tested scales, and a schedule selected at 162M transfers without retuning to 1.4B. 

Our mechanism experiments identify one way this delayed effect can be carried after removal. Early decay leaves a persistent weight-norm deficit and correspondingly larger relative steps; an approximate recurrence explains their slow convergence, and fixed-norm replays show that the measured relative-step trajectory is sufficient to reproduce the performance ordering and a similar loss difference. A separate Gram intervention exhibits related timing, norm, and post-removal relative-step behavior. These results explain how an early intervention can continue to alter optimization after it is removed, but not why a particular relative-step trajectory is beneficial: larger relative steps are not universally better, and we do not establish that norms are the only persistent state.

The practical implication is methodological: temporary schedules should be evaluated at the training budget at which the model will be used, not only when the intervention ends. Our conclusions are specific to the tested regimes. Most configurations outside the replicated comparisons use a single seed (Appendix~\ref{app:setup}). The numerical decay coefficients are not universal; detailed mechanism measurements are concentrated at 162M; some robustness comparisons use constant \(0.01\) rather than a full constant-decay sweep; and early decay does not improve every tested optimizer setting. Finally, the downstream evaluations at 405M and 1.4B do not show a clear advantage over constant decay; at 3.4B, early decay improves mean zero-shot accuracy over all tested references (Appendix~\ref{app:b34}). Whether similar delayed dynamics arise broadly across optimizers and temporary interventions remains an open question.

\bibliographystyle{assets/plainnat}
\bibliography{references}

\clearpage
\beginappendix

Appendix~\ref{app:setup} gives the training and evaluation details.
Appendices~\ref{app:gram} and~\ref{app:schedules} report additional Gram and weight-decay
comparisons. Appendices~\ref{app:domains} and~\ref{app:components} contain supporting domain and
component analyses. Appendix~\ref{app:scope} covers warmup and downstream evaluation, and
Appendix~\ref{app:extra} gives the norm-dynamics checks. Appendices~\ref{app:norm-reset}, \ref{app:linear}, \ref{app:lr}, and~\ref{app:b34} give the norm-reset, gradual-decay, learning-rate, and 3.4B experiments.

\section{Training and Evaluation Details}
\label{app:setup}

\paragraph{Experiment map} 
Main timing and scale experiments use FineWeb100B,  a 100B-token sample of FineWeb, while the replicated mechanism experiments use FineWeb-Edu \citep{penedo2024the}. 
Table~\ref{tab:map} gives the dataset and reference for each experiment group. Losses are compared only within matched groups.  Domain experiments additionally use OpenWebMath
\citep{paster2023openwebmath}, GitHub-Code-Clean \citep{codeparrotgithub}, and multilingual C4
\citep{raffel2020exploring}. 

\begin{table}[!htbp]
 \centering
 \caption{\textbf{Experiment map.} Each result is compared only with the reference in the same row.}
 \label{tab:map}
 \vspace{0.55em}
 \small
 \begin{tabular}{llll}
 \toprule
 Purpose & Data & Scale & Reference \\
 \midrule
 Replicated comparison & FineWeb-Edu & 162M & No decay; best constant \\
 Timing and scale & FineWeb100B & 162M--1.4B & No decay; best constant \\
 Norm and replay controls & FineWeb-Edu & 162M & No decay \\
  Norm reset & FineWeb-Edu; FineWeb100B & 162M; 405M & No decay; early decay \\
  Gradual decay; rate tuning & FineWeb100B & 162M; 405M & No decay; best constant \\
  Larger-scale comparison & FineWeb100B & 3.4B & No decay; best constant \\
 Warmup & FineWeb-Edu; FineWeb100B & 162M--1.4B & No decay; best constant \\
 Domain comparison & Four corpora & 162M & Constant $0.01$ \\
 Gram penalty across scale & FineWeb-Edu; FineWeb100B & 162M--1.4B & Constant $0.01$ \\
 \bottomrule
 \end{tabular}
\end{table}

\paragraph{Model and optimizer.}
The models use pre-norm Transformer blocks, parameter-free RMS normalization, QK
normalization, rotary position embeddings with base 10,000, a vocabulary of 50,304, and
untied input and output embeddings. The main MLP uses a squared-ReLU activation and hidden
width $4d$, where $d$ is the model width. Table~\ref{tab:setup} lists the dimensions and
compute-optimal budgets. Training code is based on publicly available code:
\href{https://github.com/microsoft/dion}{https://github.com/microsoft/dion}.

\begin{table}[!htbp]
 \centering
 \caption{Model dimensions and compute-optimal budgets. Every update processes
 1,024 sequences of 1,024 tokens. The duration grid uses 16,000 updates at 162M.}
 \label{tab:setup}
 \begin{tabular}{lrrrr}
 \toprule
 Parameters & Width & Layers & Heads & Updates \\
 \midrule
 162M & 768 & 12 & 6 & 3,200 \\
 405M & 1,024 & 24 & 32 & 8,100 \\
 1.4B & 2,048 & 24 & 32 & 28,000 \\
  3.4B & 3,072 & 27 & 48 & 66,000 \\
 \bottomrule
 \end{tabular}
\end{table}

Muon uses learning rate 0.01, momentum 0.95, spectral-norm learning-rate adjustment, and no
gradient clipping. Embeddings and the output head use AdamW with $\beta=(0.9,0.95)$ and no
decay. The learning rate decreases linearly to zero over the final 10\% of updates, and there
is no warmup except in Appendix~\ref{app:scope}. Matrices are initialized with standard
deviation $\min(1,\sqrt{d_{\rm out}/d_{\rm in}})/\sqrt{d_{\rm in}}$, and embeddings with unit
standard deviation. Each 162M--1.4B run uses eight H100 or B200 GPUs, with gradient accumulation where needed to
reach the global batch.

\paragraph{Data and evaluation.}
All corpora use the GPT-2 tokenizer, with documents separated by its end-of-text token. The
mathematical, code, and multilingual corpora each contain 2B training tokens and a 50M-token
validation shard. Validation loss is computed on 10,485,760 tokens, usually every 100 updates.
The domain analyses instead use a fixed set of 33,554,432 validation targets for each
paired comparison, so their losses differ slightly from the training-log values.

\paragraph{Interventions.}
Weight-decay windows use the half-open interval $[s,e)$ of Equation~\ref{eq:update}. The Gram
term in Equation~\ref{eq:gram} uses the input-side Gram matrix $W^\top W$ of each target matrix
and is summed over targets and layers without normalization. In the five-seed 162M experiment,
$\mu=10^{-3}$, $e=320$, and simultaneous decay uses $\lambda=0.3$ over the same window. Relative
steps are computed from the weight change between consecutive updates and averaged over the
layers of each matrix family.

\paragraph{Seeds.}
The replicated comparisons are the five-seed 162M experiment (Table~\ref{tab:seeds}), the
three-seed 405M comparison of early decay and no decay, the four-seed warmup experiment, the
two-seed matrix-family comparison, and the three-seed fixed-norm replay. Other configurations
use a single seed. Losses are compared only within matched groups.

\section{Gram-Penalty Scale and Target Ablations}
\label{app:gram}

\paragraph{Spectral motivation.}
In initial AdamW runs, the stable rank $\|W\|_F^2/\|W\|_2^2$, which is large when many
singular values are of similar size and small when a few dominate, decreases during early
training, particularly in $\WV$, $\WO$, and $\Wdown$. This separate observation motivated the
Gram targets but does not explain the gains in the Muon experiments.

The across-scale Gram comparisons in Table~\ref{tab:gram-scale} keep constant decay of $0.01$ throughout training
and add the penalty on $\{\WV,\WO,\Wdown\}$ during the first 10\% of updates.
Table~\ref{tab:gram-scale} shows that the penalty lowers loss by 0.013 to 0.015 nats at 162M and
405M on both datasets. At 1.4B, $\mu=10^{-3}$ is too strong and $\mu=10^{-4}$ helps.

\begin{table}[!htbp]
 \centering
 \caption{Gram regularization across scale. Differences are from the constant-$0.01$
 reference, in nats. The 1.4B reference (2.6796) is a separate run of the same
 constant-$0.01$ configuration; the run in Table~\ref{tab:scale-full} reaches 2.6802.}
 \label{tab:gram-scale}
 \begin{tabular}{llrrrr}
 \toprule
 Data & Scale & $\mu$ & Reference & Gram & Difference \\
 \midrule
 FineWeb-Edu & 162M & $10^{-3}$ & 3.1829 & 3.1699 & $-0.0130$ \\
 FineWeb-Edu & 405M & $10^{-3}$ & 2.8551 & 2.8421 & $-0.0130$ \\
 FineWeb100B & 162M & $10^{-3}$ & 3.3820 & 3.3673 & $-0.0147$ \\
 FineWeb100B & 405M & $10^{-3}$ & 3.0480 & 3.0332 & $-0.0148$ \\
 FineWeb100B & 1.4B & $10^{-3}$ & 2.6796 & 2.6813 & $+0.0017$ \\
 FineWeb100B & 1.4B & $10^{-4}$ & 2.6796 & 2.6742 & $-0.0054$ \\
 \bottomrule
 \end{tabular}
\end{table}

\paragraph{Strength and targets.}
At 405M on FineWeb-Edu, every $\mu$ from $10^{-4}$ to $10^{-2}$ improves on the reference, with
gains between 0.009 and 0.013 nats. Target ablations identify $\Wdown$ as the main contributor:
$\Wdown$ alone recovers 0.0117 of the 0.0130-nat three-target gain, whereas target sets that
exclude it improve loss by at most 0.0005 nats. Penalizing all six matrix families changes loss
by $+0.0068$ nats at 405M but $-0.0071$ at 162M. SwiGLU experiments also favor $\Wdown$ over all
three targets, so the broad six-family penalty is not a reliable replacement.

\paragraph{Timing.}
The Gram penalty has the same timing pattern as weight decay. At 162M on FineWeb-Edu
(3,200 updates), a 320-update penalty starting at 0\%, 20\%, and 60\% of training changes final
loss by $-0.0147$, $-0.0080$, and $+0.0209$ nats; decay with $\lambda=0.3$ in the same windows
gives $-0.0177$, $-0.0241$, and $+0.0063$.

\section{Replication, Schedule Sweeps, and Optimizer Variants}
\label{app:schedules}

\paragraph{Replicated intervention comparison and baseline comparison.}
Table~\ref{tab:seeds} summarizes the matched 162M comparison. Each individual intervention
improves on no regularization. Combining the Gram penalty with weight decay gives a smaller
additional change than either intervention gives relative to no regularization. For ``Both'', the
seed-1234 run computes the penalty gradient directly, and the other seeds use the
radial--tangential split with both parts at full scale; the two implementations have the same
gradient.

\begin{table}[!htbp]
 \centering
\caption{Mean final loss in five matched 162M runs
(FineWeb-Edu, 3,200 updates). Temporary interventions end at
update 320. Values are mean $\pm$ sample standard deviation; paired differences are computed before rounding.
The constant coefficient $0.05$ is the best sampled from
$\{0.01,0.05,0.1,0.2,0.3\}$.}
\label{tab:seeds}
\begin{tabular}{lcc}
\toprule
Intervention & Final loss & Difference from none \\
\midrule
No regularization & $3.1873 \pm 0.0028$ & 0 \\
Constant decay ($\lambda=0.05$) & $3.1775 \pm 0.0028$ & $-0.0098$ \\
Gram penalty & $3.1732 \pm 0.0027$ & $-0.0142$ \\
Early weight decay ($\lambda=0.3$) & $3.1701 \pm 0.0027$ & $-0.0172$ \\
Both & $3.1682 \pm 0.0030$ & $-0.0191$ \\
\bottomrule
\end{tabular}
\end{table}

\paragraph{Start time.}
Figure~\ref{fig:timing} uses 10\% windows starting at 0\%, 10\%, 20\%, 40\%, and 60\% of training
at 162M, and at 0\%, 20\%, and 60\% at 405M and 1.4B. The coefficient is $\lambda=0.3$ at 162M
and 405M and $\lambda=0.2$ at 1.4B. The 405M comparison at equal continuation evaluates the three
runs at updates 3,240, 4,860, and 8,100, each 2,430 updates after removal.

\paragraph{Window length.}
In the duration grid, constant $0.05$ is the best constant baseline, improving on no decay by
0.0122 nats. For a window in the constant-learning-rate phase,
$D=-e\log(1-0.01\lambda)$; the three outlined cells in Figure~\ref{fig:duration} all have
$D\approx2.10$. Longer windows help only up to a point (Table~\ref{tab:window-length}): the gain
is largest for a window over 50\% of training at 162M and 20\% at 405M. The 75\% window hurts at
405M, and decay throughout training hurts at both scales.

\begin{table}[!htbp]
 \centering
 \caption{Final loss difference from no decay for windows starting at initialization with
 $\lambda=0.3$, by window length as a fraction of training (FineWeb-Edu; 3,200 updates at 162M
 and 8,100 at 405M). The 162M row uses seed 1235. Differences in nats; ``--'' marks lengths we
 did not run.}
 \label{tab:window-length}
 \begin{tabular}{lrrrrrr}
 \toprule
 Scale & 5\% & 10\% & 20\% & 50\% & 75\% & 100\% \\
 \midrule
 162M & -- & $-0.0177$ & $-0.0302$ & $\mathbf{-0.0374}$ & $-0.0140$ & $+0.0279$ \\
 405M & $-0.0145$ & $-0.0249$ & $\mathbf{-0.0352}$ & $-0.0283$ & $+0.0112$ & $+0.0880$ \\
 \bottomrule
 \end{tabular}
\end{table}

\paragraph{Scale.}
Table~\ref{tab:scale-full} lists the final losses behind Table~\ref{tab:scale-practical},
including every constant coefficient in the main fixed-learning-rate comparison at these scales. The transferred schedule keeps $\lambda=0.15$ and
rounds the window to 44\% of training (12,320 of 28,000 updates). It is 0.2616 nats worse than no
decay at removal and first overtakes it at update 20,440. The best sampled schedules are $\lambda=0.3$ for the first 1,400 updates at 162M,
$\lambda=0.1$ for the first 2,100 updates at 405M, and $\lambda=0.2$ from 20\% to 30\% of training
at 1.4B. At 405M we repeated no decay and the first-10\% window with seeds 1234, 1235, and 1236;
early decay improves on no decay by 0.0250, 0.0262, and 0.0247 nats. Against constant $0.02$,
the paired improvements are 0.0175, 0.0194, and 0.0172 nats. The seed-1234 values use the run in
Table~\ref{tab:scale-practical}.

\begin{table}[!htbp]
 \centering
 \small
 \caption{Final losses behind Table~\ref{tab:scale-practical} (FineWeb100B, compute-optimal
 budget, no warmup, seed 1234). The best sampled reference at each scale is in bold.
 ``--'' marks coefficients we did not run.}
 \label{tab:scale-full}
 \begin{tabular}{lrrr}
  \toprule
  Schedule & 162M & 405M & 1.4B \\
  \midrule
  No decay & 3.3860 & 3.0537 & 2.6754 \\
  Constant $0.001$ & 3.3852 & 3.0529 & 2.6744 \\
  Constant $0.003$ & -- & -- & \textbf{2.6743} \\
  Constant $0.005$ & 3.3836 & 3.0502 & -- \\
  Constant $0.01$ & 3.3820 & 3.0480 & 2.6802 \\
  Constant $0.02$ & 3.3796 & \textbf{3.0462} & -- \\
  Constant $0.05$ & \textbf{3.3768} & 3.0497 & 2.7151 \\
  Constant $0.10$ & 3.3805 & -- & 2.7558 \\
  \midrule
  First 10\% ($\lambda$) & 3.3710 (0.3) & 3.0287 (0.3) & 2.6506 (0.2) \\
  Transferred, $\lambda=0.15$ for 44\% & -- & -- & 2.6479 \\
  Best sampled schedule & 3.3510 & 3.0180 & 2.6459 \\
  \bottomrule
 \end{tabular}
\end{table}

\paragraph{Longer training.}
\label{app:horizon}
Table~\ref{tab:horizon} lists 162M runs on FineWeb100B with 64,000 updates, 20 times the
compute-optimal budget. ``Stop at reduction'' applies decay until the final learning-rate
reduction begins at update 57,600. We report the best coefficient for each schedule class.

\begin{table}[!htbp]
 \centering
 \caption{162M runs with 64,000 updates. Differences from no decay, in nats.}
 \label{tab:horizon}
 \begin{adjustbox}{max width=\linewidth}
 \begin{tabular}{lrr}
 \toprule
 Schedule & Final loss & Difference \\
 \midrule
 No decay & 3.0826 & 0 \\
 Best constant ($0.01$) & 3.0665 & $-0.0161$ \\
 Best stop at reduction ($0.05$) & 3.0617 & $-0.0209$ \\
 First 10\%, $\lambda=0.2$ & 3.0485 & $-0.0341$ \\
 \bottomrule
 \end{tabular}
 \end{adjustbox}
\end{table}

\paragraph{Other optimizers and MLPs.}
\label{app:regime}
Table~\ref{tab:regime} compares early decay with no decay in other settings. Early decay helps in
every listed setting except AdamW with $\reluq$ at 405M. The Muon and AdamW rows also differ in
learning rate, warmup, and coefficient grid, so this exception cannot be attributed to the
optimizer alone. The sampled 162M AdamW and SwiGLU runs also improve on constant $0.01$.

\begin{table}[!htbp]
 \centering
 \caption{Best sampled first-10\% window versus no decay. FineWeb-Edu at the
 compute-optimal budget, except the Muon--$\reluq$ 1.4B row (FineWeb100B, 6,000 updates). Muon
 rows use no warmup and AdamW rows use 10\% warmup. For AdamW at 405M, $\lambda$ is the
 smallest coefficient tried. Differences in nats.}
 \label{tab:regime}
 \begin{tabular}{lllrrr}
 \toprule
 Optimizer & MLP & Scale & No decay & Early $\lambda$ & Difference \\
 \midrule
 Muon & $\reluq$ & 162M & 3.1864 & 0.3 & $-0.0171$ \\
 Muon & $\reluq$ & 405M & 2.8601 & 0.3 & $-0.0249$ \\
 Muon & $\reluq$ & 1.4B & 2.8811 & 0.2 & $-0.0181$ \\
 Muon & SwiGLU & 162M & 3.2019 & 0.3 & $-0.0189$ \\
 Muon & SwiGLU & 405M & 2.8790 & 0.3 & $-0.0235$ \\
 AdamW & SwiGLU & 162M & 3.2401 & 3.0 & $-0.0168$ \\
 AdamW & $\reluq$ & 162M & 3.2190 & 3.0 & $-0.0157$ \\
 AdamW & $\reluq$ & 405M & 2.9134 & 1.5 & $+0.0109$ \\
 \bottomrule
 \end{tabular}
\end{table}

\section{Domain Comparison}
\label{app:domains}

\begin{figure}[H]
 \centering
 \includegraphics[width=0.94\linewidth]{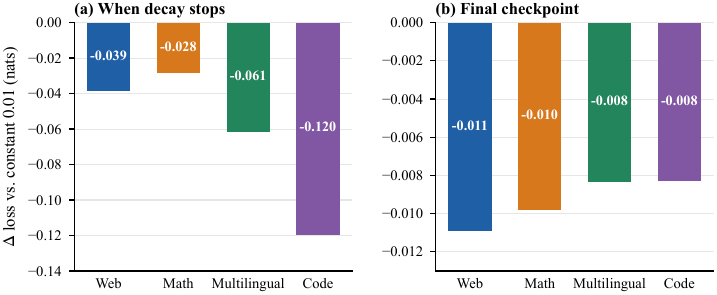}
 \caption{\textbf{The domain ranking changes between removal and the final checkpoint.}
 Validation-loss differences from constant $0.01$ when decay is removed and at
 the final checkpoint; negative values favor temporary decay. Each comparison uses a 162M
 model; temporary decay uses $\lambda=0.3$ for the first 10\% of
 training. The vertical scales differ.}
 \label{fig:domains-main}
\end{figure}

\paragraph{Data effects across domains.}
\label{sec:domains}
We compare temporary decay with constant $0.01$ in matched 162M runs on web, mathematics, multilingual text, and code. Temporary decay improves loss in all four domains both when decay is removed and at the final checkpoint, but the ranking changes substantially between the two evaluations
(Figure~\ref{fig:domains-main}; Table~\ref{tab:domain-values}). In particular, the domains with the largest immediate gains do not have the largest final gains.

\begin{table}[!htbp]
 \centering
 \caption{Domain cross-entropy and paired loss differences. Temporary decay uses
 $\lambda=0.3$ for 320 of 3,200 updates; the reference uses constant $0.01$. All values in nats.}
 \label{tab:domain-values}
 \begin{tabular}{lrrr}
 \toprule
 Domain & $H_{\rm bi}$ & $\Delta_{320}$ & $\Delta_{3200}$ \\
 \midrule
 Web & 5.982 & $-0.0388$ & $-0.0109$ \\
 Mathematics & 5.513 & $-0.0284$ & $-0.0098$ \\
 Multilingual & 4.245 & $-0.0614$ & $-0.0083$ \\
 Code & 3.815 & $-0.1196$ & $-0.0083$ \\
 \bottomrule
 \end{tabular}
\end{table}

\paragraph{Bigram cross-entropy.}
Motivated by recent work studying the emergence of token associations during early Transformer training \citep{im2026how}, we use bigram cross-entropy as a simple measure of how strongly next-token prediction can be supported by local token associations.

For each corpus, we fit unigram and bigram counts on three 100M-token training shards and
apply additive smoothing toward the training unigram distribution with strength 5. We evaluate
the resulting bigram cross-entropy $H_{\rm bi}$ on the same validation targets as the paired
model losses. This quantity measures how difficult a domain is to predict from the previous
token alone.

\paragraph{Data repetition.}
The mathematical, code, and multilingual runs make 1.68 passes over 2B-token training pools,
whereas the web runs never repeat data. We therefore repeated the web comparison on a 2B-token
FineWeb-Edu pool with the same 1.68 passes. Early decay improves on constant $0.01$ by 0.0349 nats
at removal and by 0.0128 at the final checkpoint, so web text again has a small gain at removal
and the largest final gain. Repetition therefore does not explain the web result. These losses
come from the training log, and this run is not included in
Table~\ref{tab:domain-values}.

\section{Gram Controls and Schedule Algebra}
\label{app:components}

Table~\ref{tab:components} summarizes the three controls used to separate norm and directional
effects. The following paragraphs provide the corresponding implementation details and compact
summaries of the coefficient sweeps.

\begin{table}[!htbp]
 \centering
 \small
 \caption{Gram-gradient components and copied Gram state. All runs use 162M models,
 FineWeb-Edu, 3,200 updates, and seed 1234; regularization ends at update 320. ``Before Muon''
 adds a component to the task gradient before orthogonalization. ``After Muon'' applies it
 directly to the weights. ``Copied state'' continues run A after replacing the indicated Gram
 statistics with those from penalized run B. Recovery is
 $(\mathcal L_A-\mathcal L)/(\mathcal L_A-\mathcal L_B)$. Before-Muon runs use $\mu=10^{-3}$
 and no background decay (no regularization: 3.1864). After-Muon runs use constant
 background decay $0.01$ and $\mu=0.01$ (without a penalty: 3.1829). Copied-state runs also use
 constant background decay $0.01$ but restart from saved checkpoints; their reference is the
 $A\to B$ row.
 Losses are comparable only within a block.}
 \label{tab:components}
 \begin{tabular}{llr}
 \toprule
 & Contribution & Final loss \\
 \midrule
 \multirow{5}{*}{Before Muon} & Full: $g_{\parallel}+g_{\perp}$ & 3.1718 \\
 & Norm only: $a g_{\parallel}$ & 3.1887--3.1981 \\
 & Direction only: $g_{\perp}$ & 3.1868 \\
 & Full plus early weight decay ($\lambda=0.3$) & 3.1663 \\
 & Direction plus early weight decay ($\lambda=0.3$) & 3.1710 \\
 \midrule
 \multirow{3}{*}{After Muon} & Full penalty gradient & 3.1750 \\
 & Norm only & 3.1745 \\
 & Direction only & 3.1830 \\
 \midrule
 \multirow{4}{*}{Copied state} & Normalized column correlations (27\% recovery) & 3.1833 \\
 & Column norms (62\%) & 3.1786 \\
 & Full Gram matrix (81\%) & 3.1761 \\
 & A $\rightarrow$ B endpoints & 3.1868 $\rightarrow$ 3.1736 \\
 \bottomrule
 \end{tabular}
\end{table}

\begin{figure}[!htbp]
 \centering
 \includegraphics[width=0.72\linewidth]{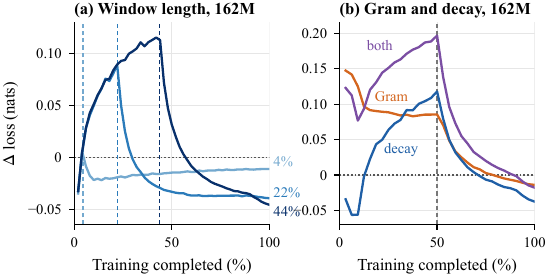}
 \caption{\textbf{Loss differences evolve after removal.}
 (a) Three 162M decay windows at $\lambda=0.15$ (FineWeb100B).
 (b) Gram regularization, weight decay, and their combination over the first 50\% at 162M
 (FineWeb-Edu). Dashed lines mark removal. References use no regularization; vertical scales
 differ.}
 \label{fig:curves}
\end{figure}

\paragraph{Gradient-component sweep.}
In the 162M experiment, adding $\mu(a g_{\parallel}+b g_{\perp})$ before Muon does not reproduce
the full penalty when either component is used alone. Across the tested coefficients, norm-only
runs end between 3.1887 and 3.1981, direction-only runs end between 3.1868 and 3.1870, and the
full penalty reaches 3.1718; no regularization reaches 3.1864. Table~\ref{tab:components}
summarizes the more informative controls that apply the components after Muon or copy the
resulting state.

\paragraph{Gram penalty combined with weight decay.}
At 162M, we keep $\mu=10^{-3}$ and $\lambda=0.3$ fixed and vary their shared window
over the first 10\%, 20\%, or 50\% of training. The combination always beats no regularization,
but its gain is smaller than the sum of the individual gains. For the 20\% and 50\% windows, it
also underperforms weight decay alone; for the 10\% window, it improves on weight decay alone
($-0.0191$ versus $-0.0177$, seed 1235). Figure~\ref{fig:curves}(b) shows the 50\% trajectories.
At 405M with a 10\% window, the combination is worse than weight decay alone with $\lambda=0.3$
(2.8455 versus 2.8352) and better with $\lambda=0.12$ (2.8364 versus 2.8413).

\paragraph{Matrix families.}
With constant $0.01$ decay kept on all matrices, adding early decay to either
$\{\WV,\WO,\Wdown\}$ or $\{\WQ,\WK,\Wup\}$ gives similar gains of 0.008 nats. Applying it to all
six families approximately adds these gains. At a matched decay budget, distributing decay over
all six families also performs better than concentrating it on the three Gram targets.

\paragraph{Gram-gradient decomposition.}
\label{app:derivations}
Write $G=W^\top W$ and $C=\OffDiag(G)$. Since $C$ is symmetric with zero diagonal,
\begin{equation}
 dE=2\langle C,dG\rangle_F
 =2\langle C,W^\top dW+(dW)^\top W\rangle_F
 =4\langle WC,dW\rangle_F.
\end{equation}
Hence $\nabla E=4WC$ and
$\langle W,\nabla E\rangle_F=4\tr(GC)=4\tr(C^2)=4E$.
Projecting onto $W$ gives, for $W\ne0$,
\begin{equation}
 \nabla E(W)=4WC
 =\underbrace{\frac{4E(W)}{\|W\|_F^2}W}_{g_{\parallel}}
 +\underbrace{\left(4WC-\frac{4E(W)}{\|W\|_F^2}W\right)}_{g_{\perp}},
 \qquad \langle W,g_{\perp}\rangle_F=0.
 \label{eq:gram-decomposition}
\end{equation}
This identity holds
for the raw gradient. It does not imply that the orthogonalized Muon update has the same
decomposition.

\paragraph{Total shrinkage.}
Let $q_t=1-\eta_t\lambda_t$ and $A_{a:b}=\prod_{j=a}^{b-1}q_j$, with
$A_{b:b}=1$. Repeated substitution in Equation~\ref{eq:update} gives
\begin{equation}
 W_T=A_{0:T}W_0-\sum_{k=0}^{T-1}\eta_k A_{k+1:T}U_k.
 \label{eq:unrolled}
\end{equation}
Matching $D$ fixes the factor $A_{0:T}=e^{-D}$ on the initial weights but not the factors on
later updates, and the updates themselves depend on the earlier trajectory. Schedules with the
same $D$ can therefore reach different parameters. The identity does not predict which schedule
reaches lower loss.

\clearpage
\section{Warmup and Downstream Results}
\label{app:scope}

\paragraph{Warmup.}
Table~\ref{tab:warmup-scale} repeats the comparison with a linear warmup over the first 10\% of
training. When the decay window is the warmup window, early decay beats the best sampled constant
at 405M and 1.4B but ties it at 162M. In the 3,200- and 8,100-update runs, all four tested windows
that extend past warmup beat the best sampled constant, by 0.011 to 0.025 nats;
Table~\ref{tab:warmup-scale} reports two representative settings.

\begin{table}[!htbp]
 \centering
 \small
 \caption{\textbf{Early decay under 10\% learning-rate warmup.} Muon, $\reluq$. ``Const.''\ is
 the best sampled constant, with its coefficient in parentheses. The first three rows use a decay
 window equal to warmup; the final two extend it. The 162M rows are means over four matched seeds;
 the 405M and 1.4B rows are single runs. FineWeb-Edu at 162M and 405M, FineWeb100B at 1.4B.
 Differences are in nats.}
 \label{tab:warmup-scale}
 \begin{tabular}{lrlrr}
  \toprule
  Scale / updates & Const. & Early ($\lambda$, window) & vs.\ none & vs.\ const. \\
  \midrule
  162M / 3,200 & 3.1944 (0.05) & 3.1940 (0.6, 320) & $-0.0094$ & $-0.0004$ \\
  405M / 8,100 & 2.8724 (0.02) & 2.8648 (0.3, 810) & $-0.0130$ & $-0.0076$ \\
  1.4B / 6,000 & 2.8965 (0.01) & 2.8847 (0.4, 600) & $-0.0129$ & $-0.0118$ \\
  \midrule
  162M / 3,200 & 3.1944 (0.05) & 3.1730 (0.3, 960) & $-0.0304$ & $-0.0214$ \\
  405M / 8,100 & 2.8724 (0.02) & 2.8473 (0.3, 1,620) & $-0.0305$ & $-0.0251$ \\
  \bottomrule
 \end{tabular}
\end{table}

With warmup and 16,000 updates at 162M (FineWeb100B), $\lambda=0.15$ over the first 7,000 updates
reaches 3.1455, which is 0.0329 nats below the best constant (0.05, 3.1784). It trails no decay by
0.0237 at update 8,000 and leads by 0.0373 at update 14,400, so warmup does not remove the delayed
crossover seen without warmup.

\paragraph{Downstream evaluation.}
Table~\ref{tab:downstream} reports zero-shot accuracy (task-normalized where applicable) for the
10\% windows; the 1.4B 44\%-window checkpoint was not kept. The 405M entries average three seeds and use
update 8,000 of 8,100. Mean accuracy differs by at most 0.11 points between early decay and
constant $0.01$ at either scale, and at 1.4B early decay is 0.65 points above no decay.

\begin{table}[!htbp]
 \centering
 \small
 \caption{Zero-shot accuracy (\%). The 405M models are trained on FineWeb-Edu and the 1.4B
 models on FineWeb100B. HS: HellaSwag; ARC-E: ARC-Easy; WG: WinoGrande.}
 \label{tab:downstream}
 \begin{tabular}{llrrrrrr}
 \toprule
 Scale & Schedule & HS & ARC-E & PIQA & LAMBADA & WG & Mean \\
 \midrule
 405M & Constant $0.01$ & 38.21 & 52.40 & 65.81 & 30.19 & 52.80 & 47.88 \\
 405M & Gram & 38.67 & 52.83 & 65.83 & 30.10 & 52.46 & 47.98 \\
 405M & Early $0.3$ & 38.72 & 52.30 & 66.21 & 30.65 & 51.85 & 47.95 \\
 \midrule
 1.4B & No decay & 54.97 & 54.59 & 73.67 & 52.59 & 56.20 & 58.40 \\
 1.4B & Constant $0.01$ & 55.78 & 54.12 & 72.52 & 53.31 & 58.96 & 58.94 \\
 1.4B & Early $0.2$ & 56.92 & 52.65 & 73.83 & 54.55 & 57.30 & 59.05 \\
 \bottomrule
 \end{tabular}
\end{table}

\clearpage
\section{Norm-Dynamics Checks}
\label{app:extra}

\paragraph{Norm dynamics after the window.}
Write the update of a weight matrix as $W_{t+1}=W_t+\Delta W_t$, where $\Delta W_t$ includes
any decay term, and let
\begin{equation}
 a_t=\frac{\|\Delta W_t\|_F}{\|W_t\|_F},\qquad
 c_t=\frac{\langle W_t,\Delta W_t\rangle_F}{\|W_t\|_F\,\|\Delta W_t\|_F}.
\end{equation}
The relative step size of Equation~\ref{eq:angular} is $\alpha_t=a_t\sqrt{1-c_t^2}$. Thus, a
recurrence for $a_t$ transfers to $\alpha_t$ only when $c_t$ is small or changes similarly across
the runs being compared. Expanding $\|W_t+\Delta W_t\|_F^2$ gives the exact identity
\begin{equation}
 \|W_{t+1}\|_F^2=\|W_t\|_F^2\,\bigl(1+2c_ta_t+a_t^2\bigr).
 \label{eq:norm-recursion}
\end{equation}
If, in addition, consecutive Muon updates have equal Frobenius norm, which Muon's normalization
makes approximately true at a constant learning rate without decay, dividing
Equation~\ref{eq:norm-recursion} by $\|\Delta W_t\|_F^2$ gives
\begin{equation}
 \frac{1}{a_{t+1}^2}-\frac{1}{a_t^2}=1+2\kappa_t,
 \qquad
 \kappa_t=\frac{c_t}{a_t}=\frac{\langle W_t,\Delta W_t\rangle_F}{\|\Delta W_t\|_F^2}.
 \label{eq:inverse-step}
\end{equation}
Without decay, $W_t=W_0+\sum_{k<t}\Delta W_k$, so
\begin{equation}
 \kappa_t=\frac{\langle W_0,\Delta W_t\rangle_F}{\|\Delta W_t\|_F^2}
 +\sum_{k<t}\frac{\langle\Delta W_k,\Delta W_t\rangle_F}{\|\Delta W_t\|_F^2}.
 \label{eq:kappa}
\end{equation}
The second term measures how well the current update aligns with all earlier updates. The
learning rate enters through the weight changes even though no separate $\eta_t$ appears in this
expression. These equations describe norms and relative steps; they do not predict loss.

\begin{figure}[!htbp]
 \centering
 \includegraphics[width=\linewidth]{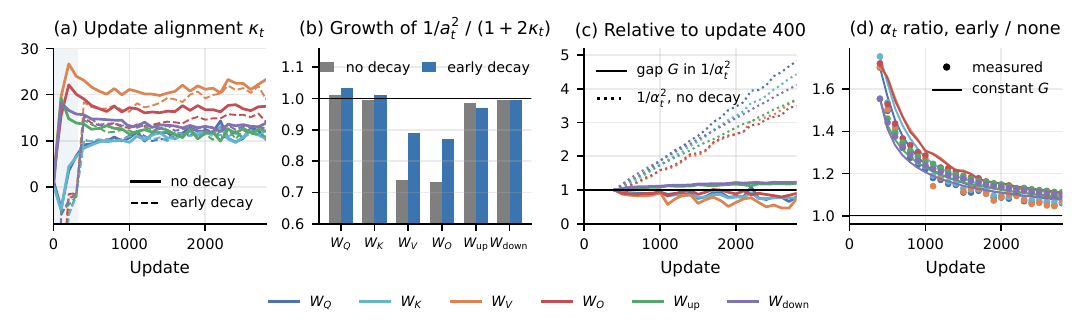}
 \caption{\textbf{Norm dynamics at 162M} (FineWeb-Edu, seed 1234): no decay versus $\lambda=0.3$ for the
 first 320 updates (shaded). (a) $\kappa_t$ of Equation~\ref{eq:inverse-step}. (b) Fitted slope
 of $1/a_t^2$ over updates 500--2,800 divided by the mean of $1+2\kappa_t$; 1 is exact agreement.
 (c) Gap $G$ in $1/\alpha_t^2$ between the runs (solid) and $1/\alpha_t^2$ of the no-decay run
 (dotted), each divided by its value at update 400. (d) Ratio of relative steps; lines hold $G$
 at its update-400 value.}
 \label{fig:norm-recursion}
\end{figure}

\paragraph{Checks at 162M.}
We test these equations on the two 162M runs of Section~\ref{sec:relative-step} (FineWeb-Edu,
3,200 updates, seed 1234), which log $\alpha_t$ and $c_t$ every 100 updates. Figure~\ref{fig:norm-recursion}
shows updates up to 2,800, before the final learning-rate reduction. The two runs have similar
$\kappa_t$ and small $|c_t|$. The recurrence closely matches norm growth for four matrix families
and overpredicts it for $\WV$ and $\WO$. The gap
$G_t=(\alpha_t^{\mathrm{none}})^{-2}-(\alpha_t^{\mathrm{early}})^{-2}$ changes much less than
$1/\alpha_t^2$: from update 400 to 2,800, $1/\alpha_t^2$ grows about fourfold, while $G_t$ varies
by up to 54\% (for $\WV$) and at update 2,800 is within 26\% of its value at update 400 in every
family. Holding $G$ at its early value follows the observed convergence of the
relative-step ratio (Figure~\ref{fig:norm-recursion}).
Over the measured interval, early decay therefore produces an approximately fixed norm deficit
when expressed in units of the update norm. Later updates grow the squared norm at similar rates
in both runs. As the norm grows, the deficit becomes a smaller fraction of it, and the relative
steps converge slowly.

\paragraph{Early learning-rate boost.}
A learning-rate boost tests the recursion with the opposite perturbation: it raises the norms
early, so later relative steps should be smaller. If the matrix learning rate is multiplied by
$r$ for the first $e$ updates and $\kappa_t$ is unchanged, $1/a_t^2$ afterwards exceeds its
no-decay value by $(r^2-1)\sum_{k<e}(1+2\kappa_k)$. We ran $r=2$ for 320 updates, $r=4$ for 320
updates, and $r=2$ for 960 updates, without decay. The offset remains approximately constant
after each boost. For $\Wup$ and $\Wdown$, it agrees with the prediction within 8\%; deviations
for the attention matrices are consistent with changes in $\kappa_t$ or unequal update norms.
The relative steps remain below the no-decay values, and all three runs end with higher loss.

\paragraph{Replaying relative steps at fixed norm.}
To isolate the relative-step trajectory, we remove norm changes. After each Muon step, a variant
of Hyperball \citep{wen2026hyperball} rescales the weight change of each matrix to Frobenius
norm $\eta_t\|W_0\|_F$ and projects the weights back to norm $\|W_0\|_F$. Weight norms then stay
at their initial values, and $\eta_t$ sets the relative step. We prescribe the relative step using the six-family mean of the logged $\alpha_t$ in either the no-decay run or the $\lambda=0.3$ run, interpolated
linearly between logged updates (Figure~\ref{fig:replay}a). The replays use 162M, FineWeb-Edu,
3,200 updates, and no weight decay. Both trajectories come from seed 1234 and are reused for all
three replay seeds.
The early-decay trajectory beats the no-decay trajectory by 0.0221--0.0233 nats across the three
replays, compared with 0.0171 nats between the Muon source runs
(Figure~\ref{fig:replay}b). Splitting the trajectories at update 320 attributes 0.0181 nats of the
0.0221-nat replay difference to the post-window steps. Replaying only the early-window trajectory
before continuing with ordinary Muon changes final loss by just 0.0003 nats.
With norms held fixed, the early-decay trajectory reproduces the ordering and approximate size
of the gain, and most of the replay gain comes from the larger relative steps after the window.
Figure~\ref{fig:norm-recursion} relates these larger steps to the reduced norms.

\paragraph{The Gram penalty raises the relative steps of the matrices it targets.}
The Gram runs also log $\alpha_t$. We divide $\alpha_t$ in each Gram run by $\alpha_t$ in a
reference with the same background decay and no penalty. Immediately after removal, the targeted
families take 1.3--2.7 times the reference relative step; near the end of training, the ratio is
1.04--1.11. After removal, untargeted families stay within 5\% of the reference when $\WV$, $\WO$,
and $\Wdown$ are all targeted. When only $\Wdown$ is targeted (SwiGLU), $\WV$ and $\Wup$ differ
from the reference by up to 10\% until about update 1,600. The Gram penalty therefore
produces the same qualitative pattern as early decay---smaller norms and larger later relative
steps---mainly in the matrices it targets.

\clearpage
\section{Norm Resets at Removal}
\label{app:norm-reset}

\paragraph{Setup.}
Do the norms left by temporary decay contribute to its later benefit in ordinary Muon?
We train a no-decay run A and an early-decay run B, with $\lambda=0.3$ over the first 10\%.
The 162M comparison uses FineWeb-Edu, 3,200 updates, and removal at 320; the 405M comparison
uses FineWeb100B, 8,100 updates, and removal at 810. Each source checkpoint is continued
unchanged and with its block matrices rescaled to the counterpart's norms. At 405M, the
no-decay source run stopped at update 5,700 after a hardware fault; its update-810 checkpoint
supplies the reset, and a rerun of the same configuration supplies the unchanged continuation.

For each recipient matrix $W$ and counterpart $V$, the reset is
\[
 W\leftarrow W\,\frac{\|V\|_F}{\|W\|_F}.
\]
It applies independently to the six matrix families in every Transformer block; embeddings
and the output head are unchanged. Directions and optimizer state are preserved at the reset.
Every continuation uses its original data order and learning-rate schedule, without decay.
These controls use their own matched source runs, and the 162M runs use FineWeb-Edu rather than
FineWeb100B, so their final losses differ from those in Table~\ref{tab:scale-practical}.

\begin{table}[!htbp]
 \centering
 
 \small
 \caption{\textbf{Final losses after norm resets.} A: no decay; B: early decay.
 A with B norms continues the no-decay checkpoint after adopting early-decay norms,
 and conversely for B with A norms. Each row is a separate matched comparison.}
 \label{tab:norm-reset-full}
 \begin{tabular}{lrrrr}
  \toprule
  Scale & A unchanged & B unchanged & A with B norms & B with A norms \\
  \midrule
  162M & 3.1866 & 3.1694 & 3.1701 & 3.1861 \\
  405M & 3.0539 & 3.0302 & 3.0323 & 3.0471 \\
  \bottomrule
 \end{tabular}
\end{table}

For $g=L_A-L_B$, recovered gain is $(L_A-L_{A\leftarrow B})/g$ and removed gain is
$(L_{B\leftarrow A}-L_B)/g$. These give 96\% and 97\% at 162M, and 91\% and 71\% at 405M.
The intervention changes model outputs as well as subsequent optimization. At the first
logged evaluation after the reset (updates 400 and 900), giving A smaller norms increases
loss by 0.062 and 0.174 nats, respectively, relative to its unchanged continuation. It
achieves lower loss only with further training. Norms therefore contribute to the subsequent
gain, although these controls do not predict its size or explain the remaining difference.

\section{Gradual Weight-Decay Schedules}
\label{app:linear}

\paragraph{Setup.}
Does the benefit require decay to stop abruptly? We compare a first-10\% window with decay
that decreases linearly to zero over the first 20\% or over the whole run:
\[
 \lambda_t=\lambda_0\max(1-t/e,0).
\]
Both scales use FineWeb100B, no warmup, the standard learning-rate schedule, and the same
training budget as their references. At 162M, $T=3,200$; at 405M, $T=8,100$.
The window and first-20\% linear schedule start at $\lambda_0=0.3$.
The full-training linear schedules start at 0.1 and 0.04, respectively. This compares tested
schedules, rather than isolating duration at equal total shrinkage.

\begin{table}[!htbp]
 \centering
 
 \small
 \caption{\textbf{Final loss for gradual schedules.} Each row uses its own reference runs.
 The 162M reference runs differ from those in Table~\ref{tab:scale-practical}.}
 \label{tab:linear-schedules}
 \begin{tabular}{lrrrrr}
  \toprule
  Scale & No decay & Best constant & First 10\% & Linear 20\% & Linear full run \\
  \midrule
  162M & 3.3869 & 3.3779 & 3.3708 & 3.3689 & 3.3668 \\
  405M & 3.0537 & 3.0462 & 3.0287 & 3.0257 & 3.0340 \\
  \bottomrule
 \end{tabular}
\end{table}

Both linear schedules improve on the best constant at each scale. The first-20\% schedule
also beats the short window at both scales, while full-training linear decay does so only at
162M. The benefit therefore extends to gradual decay, and the early window is not generally
the best schedule. Its 162M gain over the constant is 0.0071 nats in this matched group,
compared with 0.0058 in Table~\ref{tab:scale-practical}; the reference runs differ.

\section{Learning-Rate Sensitivity}
\label{app:lr}

\paragraph{Setup.}
Is the early-decay advantage caused by an unsuitable base learning rate?
The base rate of 0.01 follows \citet{ahn2025dion}, Appendix E.1.
We test base rates in $\{0.005,0.01,0.02,0.04\}$ on FineWeb100B, at the compute-optimal
budgets, without warmup. We select the no-decay and constant references separately from
their tested configurations. At 162M, we combine a rate sweep at constant decay 0.01
with a coefficient sweep at base rate 0.01 and a run with constant decay 0.0125 at rate 0.04.
At 405M, we include constants 0.005, 0.01, and 0.02 at rate 0.02 alongside the
coefficient sweep at base rate 0.01.

For the reported early schedules, changing the base rate from $\eta_0=0.01$ to $\eta$ changes
$\lambda$ from 0.3 to $0.3\eta_0/\eta$. The window remains the first 10\%.
Since the learning-rate profile is scaled by the same factor, this preserves every
$(1-\eta_t\lambda_t)$ and the total log-shrinkage $D$.
The embedding base rate follows the swept rate, and the output-head base rate is that
rate divided by $\sqrt d$.

\begin{table}[!htbp]
 \centering
 
 \small
 \caption{\textbf{Selected final losses after learning-rate tuning.}
 Parentheses give (base learning rate, decay coefficient). For early decay, the coefficient
 applies only to the first 10\%. References are selected separately.}
 \label{tab:lr-sensitivity}
 \begin{tabular}{lrrr}
  \toprule
  Scale & No decay & Constant decay & Early decay \\
  \midrule
  162M & 3.3089 (0.04, 0) & 3.3066 (0.04, 0.01) & 3.2999 (0.04, 0.075) \\
  405M & 3.0118 (0.04, 0) & 3.0170 (0.02, 0.005) & 3.0019 (0.02, 0.15) \\
  \bottomrule
 \end{tabular}
\end{table}

Early decay improves over the selected no-decay and constant references by 0.0090 and
0.0067 nats at 162M, and 0.0099 and 0.0151 at 405M. These are comparisons among tested
configurations, not claims of globally optimal rates or equivalence to a time-varying
learning-rate schedule.

\begingroup
\color{black}
\paragraph{Fixed non-matrix Adam base rate.}
Following \citet{ahn2025dion}, Appendix E.2, a separate 162M check fixes embedding and
output-head Adam base rates at 0.002 while changing the Muon rate. At Muon rate 0.04,
$\lambda=0.075$ over the first 320 updates reaches 3.3410, versus 3.3460 without decay.
The learning-rate decay over the final 10\% applies to all parameter groups.
The window has the same total multiplicative shrinkage as $\lambda=0.3$ at base rate 0.01.
\par\endgroup

\clearpage
\section{Experiments at 3.4B}
\label{app:b34}

\paragraph{Setup.}
We test whether the benefits extend to a larger model with width 3,072, 27 layers,
48 heads, and 3.37B parameters.
It trains on FineWeb100B for 66,000 updates, about 69B tokens, with the model, batch,
optimizer, and learning-rate conventions of Appendix~\ref{app:setup}. Validation is
evaluated every 200 updates. Five configurations use 32 H100 GPUs; the
2,800- and 12,320-update configurations use B200 GPUs on GKE.

We compare no decay, constants 0.001 and 0.003 (the best sampled coefficients at 1.4B), and
four early schedules. Two use $\lambda=0.2$: the first 10\% (6,600 updates), or 2,800 updates,
which preserves the length of the 1.4B first-10\% window. Two use the coefficient selected at
162M, $\lambda=0.15$: the first 44\% (29,040 updates), or 12,320 updates, which preserves the
length used in the 1.4B transfer. The transfer rules differ in duration and total shrinkage.

\begin{table}[!htbp]
 \centering
 
 \small
 \caption{\textbf{Final validation loss and mean zero-shot accuracy at 3.4B.}
 Differences are from no decay. Accuracy averages the same five tasks used below.}
 \label{tab:b34-loss}
 \begin{adjustbox}{max width=\linewidth}
 \begin{tabular}{lrrr}
  \toprule
  Schedule & Final loss & Difference & Mean accuracy (\%) \\
  \midrule
  No decay & 2.4832 & 0 & 65.30 \\
  Constant 0.001 & 2.4843 & $+0.0011$ & 65.68 \\
  Constant 0.003 & 2.4895 & $+0.0063$ & 65.69 \\
  $\lambda=0.2$, first 10\% & 2.4720 & $-0.0112$ & 66.54 \\
  $\lambda=0.2$, 2,800 updates & 2.4781 & $-0.0051$ & 65.96 \\
  Transfer: $\lambda=0.15$, first 44\% & 2.4782 & $-0.0050$ & 66.85 \\
  Transfer: $\lambda=0.15$, 12,320 updates & 2.4638 & $-0.0194$ & 66.90 \\
  \bottomrule
 \end{tabular}
 \end{adjustbox}
\end{table}

Neither constant reduces validation loss, although both increase mean accuracy.
Every early schedule improves both metrics over all three references.
The fixed-update transfer has the lowest final loss among these schedules.
These runs support transfer under the tested rules; they do not determine an optimal duration.

\begin{figure}[!htbp]
 \centering
 
 \includegraphics[width=\linewidth]{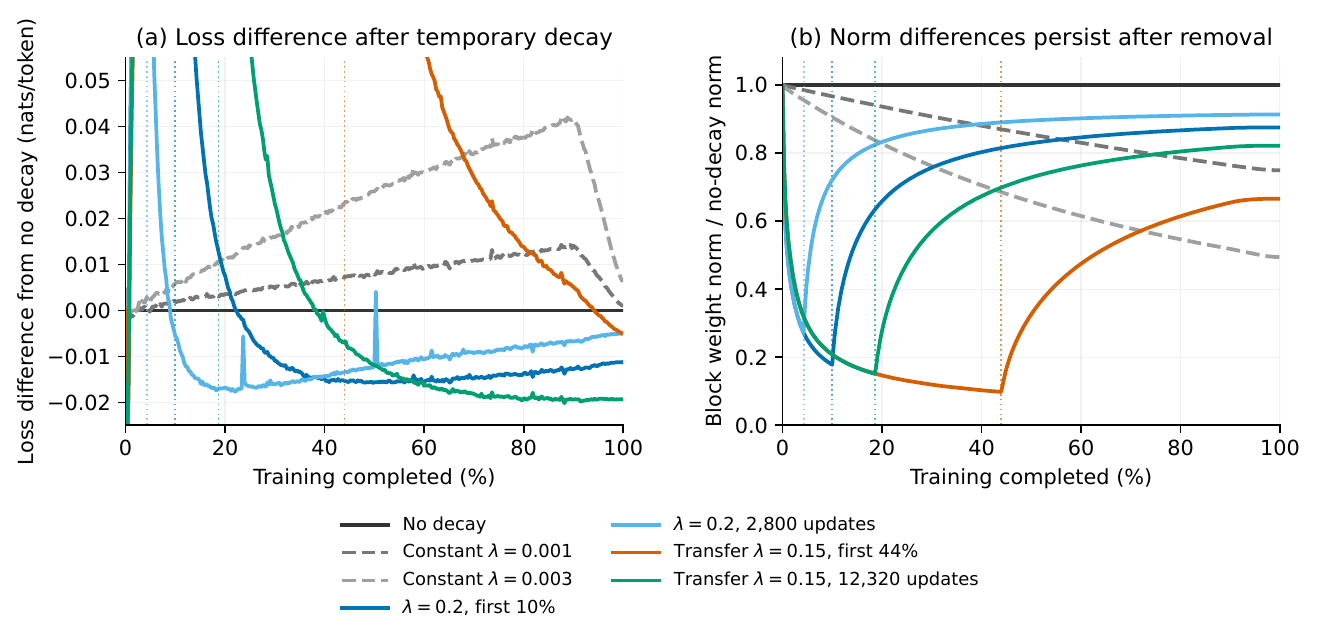}
 \caption{\textbf{Temporary decay at 3.4B.} (a) Validation-loss difference from no decay.
 Dotted vertical lines mark the end of each window. The positive range is truncated to show the
 late comparison; the removal-time differences are stated above.
 (b) Root-sum-square of the six logged matrix-family norms, divided by the corresponding no-decay quantity. All four early schedules
 and both constant references are shown.}
 \label{fig:b34}
\end{figure}

\paragraph{Delayed effects and norms.}
The first-10\% window is 0.2847 nats worse than no decay at its last active-window evaluation
and stays below the reference from update 14,800 onward. The 44\% transfer is 0.3852 nats
worse at its last active-window evaluation (update 29,000), and stays below the reference
only from update 62,200, about 94\% of training.
The 12,320-update transfer is 0.2973 nats worse at update 12,200 and stays below the reference
from update 25,400; the 2,800-update window is 0.1955 nats worse at update 2,800.
Figure~\ref{fig:b34} shows all four windows and the references. Norm ratios show persistent
differences after decay ends, but these measurements are not a direct test of the relative-step
recurrence.

\paragraph{Downstream evaluation.}
Table~\ref{tab:b34-downstream} reports accuracy on each task, using the same zero-shot metrics
as Table~\ref{tab:downstream}. Mean accuracy is supporting evidence: better validation loss
does not imply an improvement on every task, and the two transfer rules have nearly equal
mean accuracy despite different final losses.

\begin{table}[!htbp]
 \centering
 
 \small
 \caption{\textbf{Zero-shot accuracy (\%) at 3.4B.} HS: HellaSwag; ARC-E: ARC-Easy;
 WG: WinoGrande. Means are reported in Table~\ref{tab:b34-loss}.}
 \label{tab:b34-downstream}
 \begin{adjustbox}{max width=\linewidth}
 \begin{tabular}{lrrrrr}
  \toprule
  Schedule & HS & ARC-E & PIQA & LAMBADA & WG \\
  \midrule
  No decay & 65.66 & 59.97 & 77.69 & 61.13 & 62.04 \\
  Constant 0.001 & 66.34 & 61.53 & 77.48 & 61.23 & 61.80 \\
  Constant 0.003 & 66.00 & 60.31 & 77.86 & 61.83 & 62.43 \\
  $\lambda=0.2$, first 10\% & 66.95 & 61.87 & 77.75 & 62.84 & 63.30 \\
  $\lambda=0.2$, 2,800 updates & 66.71 & 60.40 & 77.69 & 61.87 & 63.14 \\
  Transfer: $\lambda=0.15$, first 44\% & 67.54 & 62.63 & 77.80 & 63.15 & 63.14 \\
  Transfer: $\lambda=0.15$, 12,320 updates & 67.72 & 61.83 & 77.80 & 63.92 & 63.22 \\
  \bottomrule
 \end{tabular}
 \end{adjustbox}
\end{table}

\end{document}